\documentclass[sigconf,natbib=true]{acmart}
\AtBeginDocument{%
  }

\setcopyright{acmlicensed}
\copyrightyear{2006}
\acmYear{2006}
\acmDOI{XXXXXXX.XXXXXXX}

\acmConference[xxx '17]{The xxth }
  {November 11--24, 2006}{xxx, xxxx}
\acmBooktitle{Proceedings of the xxx, July 11--15, 2012, xxx, xxxx}

\newcommand{\ie}{\emph{i.e., }}
\newcommand{\eg}{\emph{e.g., }}

\usepackage{arydshln}
\usepackage{booktabs}
\usepackage{multirow}
\usepackage{makecell}
\usepackage{tabularx}
\usepackage{threeparttable}
\usepackage{adjustbox}
\usepackage{pifont}
\usepackage[table]{xcolor}
\usepackage{xcolor}
\usepackage{array}
\usepackage{enumitem}

\definecolor{tablegray}{RGB}{245,246,247}
\definecolor{groupgray}{RGB}{235,237,240}
\definecolor{oursorange}{RGB}{255,244,230}

\definecolor{f1green}{HTML}{E5EFDB}
\definecolor{f1purple}{HTML}{EAE3F1}
\definecolor{f1blue}{RGB}{223,234,246}

\newcommand{\diamondbullet}{\ding{117}} 
\newcommand{\cmark}{\ding{51}}   
\newcommand{\xmark}{\ding{55}}   
\newcommand{\pmark}{\raisebox{0.2ex}{\small$\triangle$}} 

\newcommand{\datasetname}{FashionStylist}

\newcommand{\capbg}[2]{%
  \begingroup
  \setlength{\fboxsep}{1pt}%
  \raisebox{0pt}[0pt][0pt]{\colorbox{#1}{#2}}%
  \endgroup
}
\begin{document}

\title{FashionStylist: An Expert Knowledge-enhanced Multimodal Dataset for Fashion Understanding}

\author{Kaidong~Feng}
\affiliation{%
  \institution{Yanshan University}
  \city{Qinhuangdao}
  \country{China}
}
\email{kaidong3762@gmail.com}

\author{Zhuoxuan~Huang}
\affiliation{%
  \institution{Central South University}
  \city{Changsha}
  \country{China}}
\email{lilhzx@csu.edu.cn}

\author{Huizhong~Guo}
\affiliation{%
  \institution{Zhejiang University}
  \city{Hangzhou}
  \country{China}}
\email{huiz_g@zju.edu.cn}

\author{Yuting~Jin, Xinyu Chen \\ Yue~Liang, Yifei Gai \\ Li Zhou}
\affiliation{%
  \institution{Southwest University}
  \city{Chongqing}
  \country{China}}
\email{fashion@swu.edu.cn}

\author{Yunshan~Ma}
\affiliation{%
  \institution{Singapore Management University}
  \country{Singapore}
}
\email{ysma@smu.edu.sg}

\author{Zhu~Sun}
\affiliation{%
  \institution{Singapore University of Technology and Design}
  \country{Singapore}
  }
\email{zhu_sun@sutd.edu.sg}

\renewcommand{\shortauthors}{Feng et al.}

\begin{abstract}

Fashion understanding requires both visual perception and expert-level reasoning about style, occasion, compatibility, and outfit rationale. However, existing fashion datasets remain fragmented and task-specific, often focusing on item attributes, outfit co-occurrence, or weak textual supervision, and thus provide limited support for holistic outfit understanding. In this paper, we introduce FashionStylist, an expert-annotated benchmark for holistic and expert-level fashion understanding. Constructed through a dedicated fashion-expert annotation pipeline, FashionStylist provides professionally grounded annotations at both the item and outfit levels. It supports three representative tasks: outfit-to-item grounding, outfit completion, and outfit evaluation. These tasks cover realistic item recovery from complex outfits with layering and accessories, compatibility-aware composition beyond co-occurrence matching, and expert-level assessment of style, season, occasion, and overall coherence. Experimental results show that FashionStylist serves not only as a unified benchmark for multiple fashion tasks, but also as an effective training resource for improving grounding, completion, and outfit-level semantic evaluation in MLLM-based fashion systems. 
\end{abstract}



\maketitle

\section{Introduction}
Fashion is a major application domain for multimedia research, given its broad influence on consumer choice and fashion design practice~\cite{cheng2021fashion,ding2023computational,shi2025generative,su2021complementary}. 
Unlike object recognition problems that focus on isolated items, fashion understanding requires joint modeling of visual perception and semantic understanding over complete looks~\cite{ding2023computational,ma2020knowledge}. 
Users and designers typically reason about \emph{complete outfits}: they identify constituent pieces from an overall look, compose outfits from partial inputs, and assess whether a look matches a particular style, season, or occasion. These demands require models to move beyond low-level garment recognition toward holistic understanding of outfits and fashion semantics~\cite{liao2018knowledge,zhao2024unifashion,jin2025genwardrobe}.

However, existing fashion datasets still provide only limited support for this broader form of holistic fashion understanding. 
Much of the existing literature is \emph{item-centric}~\cite{liu2016DeepFashion,ge2019deepfashion2,jia2020fashionpedia, wu2021fashion, guan2022personalized}, where the basic data unit is an individual fashion item and supervision is centered on single-garment perception, such as recognition, retrieval, and virtual try-on/off. 
While valuable for fine-grained garment understanding, such datasets \textbf{\textit{provide limited support for modeling complete outfits as coherent semantic units}}.
Although some datasets include full-look images or paired outfit information, such signals are often used only as auxiliary context for individual garments, and the covered dressing scenarios are usually simplified to regular combinations. 
As a result, they rarely capture realistic outfit complexity, including accessories, layering, richer multi-item coordination, or outfit-level semantics such as occasion suitability.

\begin{table*}[t]
\centering
\footnotesize
\setlength{\abovecaptionskip}{2pt}
\setlength{\tabcolsep}{2pt}
\renewcommand{\arraystretch}{1.16}
\caption{Summary of representative fashion datasets. Existing benchmarks typically provide either item-level signals or outfit/set-level composition information, but rarely combine them with structured expert annotations for complete outfits.}
\label{tab:finegrained_annotation_comparison}
\begin{threeparttable}
\begin{adjustbox}{max width=0.95\textwidth}
\begin{tabular}{p{2.7cm}p{2.2cm}p{3.0cm}ccccccccc}
\toprule
\multirow{3}{*}{\textbf{Dataset Family}}
& \multirow{3}{*}{\textbf{Dataset}}
& \multirow{3}{*}{\textbf{Main Task}}
& \multicolumn{4}{c}{\textbf{Item-level}}
& \multicolumn{5}{c}{\textbf{Outfit-level}} \\
\cmidrule(lr){4-7} \cmidrule(lr){8-12}
& &
& \multirow{2}{*}{\makecell[c]{\textbf{Item}\\ \textbf{Image}}}
& \multirow{2}{*}{\makecell[c]{\textbf{Title/}\\ \textbf{Category}}}
& \multirow{2}{*}{\makecell[c]{\textbf{Description}}}
& \multirow{2}{*}{\makecell[c]{\textbf{Expert}\\ \textbf{Item Ann.}}}
& \multirow{2}{*}{\makecell[c]{\textbf{Full-look}\\ \textbf{Image}}}
& \multirow{2}{*}{\makecell[c]{\textbf{Outfit}\\ \textbf{Bundle/Set}}}
& \multicolumn{3}{c}{\textbf{Expert Outfit Ann.}} \\
\cmidrule(lr){10-12}
& &
& 
& 
& 
& 
& 
& 
& \textbf{Style}
& \textbf{Season}
& \textbf{Occasion} \\
\midrule
\multirow{7}{*}{\makecell[l]{Item-centric\\visual/multimodal\\understanding}}
& DeepFashion~\cite{liu2016DeepFashion}
& \makecell[l]{Fashion Recognition}
& \cmark & \pmark & \xmark & \pmark & \xmark & \xmark & \xmark & \xmark & \xmark \\
& DeepFashion2~\cite{ge2019deepfashion2}
& \makecell[l]{Dense Understanding}
& \cmark & \pmark & \xmark & \pmark & \pmark & \xmark & \xmark & \xmark & \xmark \\
& Fashionpedia~\cite{jia2020fashionpedia}
& \makecell[l]{Fashion Parsing}
& \xmark & \pmark & \xmark & \pmark & \cmark & \xmark & \xmark & \xmark & \xmark \\
& Fashion-Gen~\cite{rostamzadeh2018fashion}
& \makecell[l]{Multimodal Generation}
& \cmark & \pmark & \cmark & \pmark & \cmark & \xmark & \xmark & \xmark & \xmark \\
& FashionIQ~\cite{wu2021fashion}
& \makecell[l]{Fashion Retrieval}
& \cmark & \cmark & \cmark & \xmark & \xmark & \xmark & \xmark & \xmark & \xmark \\
& VITON-HD~\cite{choi2021viton}
& \makecell[l]{Virtual Try-on/off}
& \cmark & \pmark & \xmark & \xmark & \cmark & \xmark & \xmark & \xmark & \xmark \\
& DressCode~\cite{morelli2022dress}
& \makecell[l]{Virtual Try-on/off}
& \cmark & \pmark & \xmark & \xmark & \cmark & \xmark & \xmark & \xmark & \xmark \\
\midrule
\multirow{4}{*}{\makecell[l]{Outfit-centric\\compatibility and\\recommendation}}
& Polyvore-U~\cite{lu2019learning}
& \makecell[l]{Outfit Recommendation}
& \cmark & \cmark & \xmark & \xmark & \xmark & \cmark & \xmark & \xmark & \xmark \\
& iFashion~\cite{chen2019pog}
& \makecell[l]{Outfit Recommendation}
& \cmark & \cmark & \xmark & \xmark & \xmark & \cmark & \xmark & \xmark & \xmark \\
& IQON3000~\cite{song2019gp}
& \makecell[l]{Outfit Compatibility}
& \cmark & \cmark & \cmark & \xmark & \xmark & \cmark & \xmark & \xmark & \xmark \\
& FLORA~\cite{deshmukh2024dressing}
& \makecell[l]{Generation / Grounding}
& \xmark & \xmark & \xmark & \xmark & \cmark & \xmark & \pmark & \xmark & \xmark \\
\midrule
\rowcolor{oursorange}
\multirow{1}{*}{\textbf{Our dataset}}
& \textbf{\datasetname}
& \makecell[l]{Knowledge-aware Fashion\\Understanding}
& \textbf{\cmark} & \textbf{\cmark} & \textbf{\cmark} & \textbf{\cmark}
& \textbf{\cmark} & \textbf{\cmark} & \textbf{\cmark} & \textbf{\cmark} & \textbf{\cmark} \\
\bottomrule
\end{tabular}
\end{adjustbox}
\begin{tablenotes}
\footnotesize
\item \textbf{\cmark}: explicitly supported; \textbf{\xmark}: not supported; \textbf{\pmark}: partially or indirectly supported. ``Full-look Image'' denotes a real image showing a complete outfit worn by a person or model.
\item ``Outfit Bundle/Set'' denotes an outfit represented as a coordinated set of multiple fashion items, without requiring a real full-look worn image.
\item ``Expert Item Ann.'' denotes professional or expert-informed annotations on individual fashion items, such as gender, style, material, color, pattern, and layering role.
\end{tablenotes}
\end{threeparttable}
\end{table*}


\emph{Outfit-centric datasets}~\cite{chen2019pog,lu2019learning,song2019gp,zheng2021personalized} take the outfit, or a coordinated bundle of items, as the basic data unit, and thus move closer to realistic fashion applications. 
They typically contain outfit composition information, category metadata, user interaction signals, or weak textual descriptions, which makes them suitable for tasks such as recommendation and compatibility prediction.
However, they \textit{\textbf{provide limited support for understanding why an outfit is appropriate, what it expresses, or how it should be evaluated from a human perspective}}, because they rarely provide structured expert-level outfit semantics, such as style identity, season suitability, occasion appropriateness, or overall styling rationale.

This gap becomes most evident in real-world fashion applications. Users may wish to identify constituent items in a complete look, especially in scenarios involving accessories, layered garments, or partial occlusion. Stylists or recommender systems may need to complete an outfit from partial inputs while preserving visual compatibility, style consistency, and contextual appropriateness. More importantly, users and designers often expect systems to provide semantic judgments about a full outfit, such as style, season fit, occasion suitability, or overall coherence. These needs naturally motivate three representative tasks: \emph{outfit-to-item grounding}, \emph{outfit completion}, and \emph{outfit evaluation}. While the first two are only partially supported by existing benchmarks, the third remains largely unsupported because current datasets rarely provide the outfit-level expert annotations required for systematic fashion evaluation.

To address these limitations, we present \datasetname, a fashion benchmark developed with a dedicated \textbf{fashion expert team}. 
Unlike datasets that focus mainly on item categories, or co-occurrence signals, \datasetname\ provides professionally grounded annotations at both the \emph{item} and \emph{outfit} levels. 
It offers several key benefits. 
\textit{\textbf{First, it supports more realistic and challenging fashion scenarios.}} 
For \emph{outfit-to-item grounding}, \datasetname\ includes outfits that better reflect real dressing complexity, including accessories, layering, and richer multi-item composition, thereby making evaluation substantially closer to real-world fashion understanding. 
\textit{\textbf{Second, it provides richer semantic knowledge for reasoning about item--outfit relations}}. 
For \emph{outfit completion}, the combination of item-side and outfit-side expert annotations enables models to learn not only whether items co-occur, but also whether they are semantically compatible in terms of style, structure, and usage context. 
Beyond these two tasks, \textit{\textbf{\datasetname\ further supports \emph{outfit evaluation}}}, a task that requires expert-level judgment over complete outfits and enables direct assessment of whether current LLMs/MLLMs truly understand fashion semantics rather than merely modeling surface appearance patterns.

We evaluate \datasetname\ on these three tasks to demonstrate its value as both a benchmark and a training resource. The results show that current large models still fall short of expert-level fashion understanding, especially when reasoning about full outfits at a semantic level. In contrast, models adapted to \datasetname\ obtain clear improvements across grounding, completion, and evaluation, indicating that the dataset provides structured knowledge missing from existing fashion resources. These findings suggest that progress in fashion intelligence depends not only on larger models, but also on datasets that bridge low-level visual perception and high-level expert semantics.

The main contributions of this work are summarized as follows:
\begin{itemize}[leftmargin=2em, topsep=2pt, itemsep=1pt, parsep=0pt, partopsep=0pt]
    \item We present \datasetname, a publicly released\footnote{\url{https://github.com/recsys-benchmark/FashionStylist}}, expert-annotated fashion benchmark with professionally grounded annotations at both the item and outfit levels.
    \item We establish three benchmark tasks: \emph{outfit-to-item grounding}, \emph{outfit completion}, and \emph{outfit evaluation}, which together reflect realistic user and designer needs, while highlighting capabilities that existing fashion datasets do not adequately support.
    \item We show through extensive experiments that \datasetname\ provides valuable supervision for higher-level fashion understanding, improves model performance across multiple tasks, and enables systematic evaluation of expert-level outfit reasoning in LLMs and MLLMs.
\end{itemize}

\begin{figure*}[tbp]
    \centering
    \includegraphics[width=0.9\textwidth]{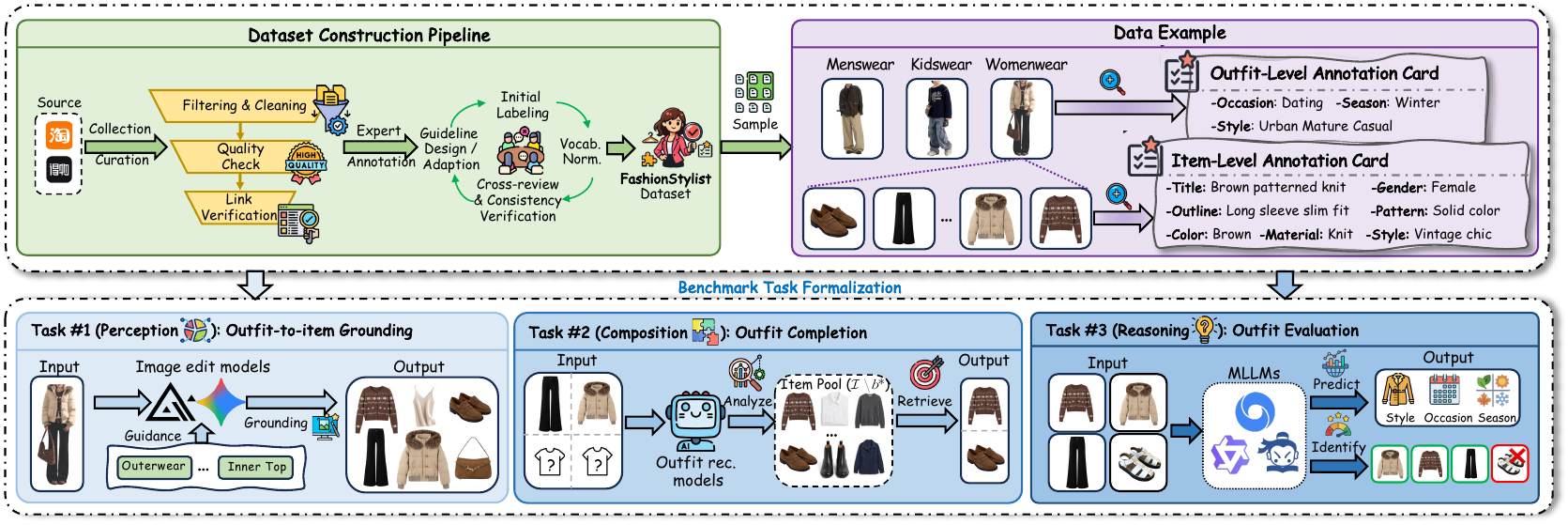}
    \caption{Overview of our proposed \datasetname, where the \capbg{f1green}{green} part presents the pipeline of dataset construction, the \capbg{f1purple}{purple} part provides toy data examples in \datasetname, and the \capbg{f1blue}{blue} part introduces the proposed benchmark tasks.}
    \label{fig:framework}
    \vspace{-0.3cm}
\end{figure*}
\section{Related Work}

Fashion datasets have supported a broad range of multimedia tasks. Rather than exhaustively reviewing all existing resources, we summarize representative benchmarks in Table~\ref{tab:finegrained_annotation_comparison} according to the supervision and task support they provide.

\emph{Item-level fashion datasets} have substantially advanced fine-grained garment perception. DeepFashion~\cite{liu2016DeepFashion}, DeepFashion2~\cite{ge2019deepfashion2}, and Fashionpedia~\cite{jia2020fashionpedia} support recognition, parsing, detection, and attribute prediction, and have become standard resources for learning visual representations of individual fashion items. As shown in Table~\ref{tab:finegrained_annotation_comparison}, these datasets provide strong item-level perception signals, but only limited outfit-level semantics. Even when full-look images are available, they are mainly used as context for item-level understanding rather than structured supervision for complete outfits. A related line of work enriches item-centric data with \emph{textual or generative supervision}. Fashion-Gen~\cite{rostamzadeh2018fashion} provides product images paired with stylist-written descriptions, while FashionIQ~\cite{wu2021fashion} supports language-guided retrieval through relative captions. Virtual try-on/off datasets, such as VITON-HD~\cite{choi2021viton} and   DressCode~\cite{morelli2022dress}, further extend this line to garment synthesis and appearance transfer. These resources broaden item-level supervision, but still do not provide structured expert-level annotations for complete outfits.

\emph{Outfit-level fashion datasets} take outfit sets or coordinated bundles as the basic unit, and thus move closer to realistic fashion applications. Polyvore-U~\cite{lu2019learning}, iFashion~\cite{chen2019pog}, and IQON3000~\cite{song2019gp} support recommendation, compatibility prediction, and personalization, while FLORA~\cite{deshmukh2024dressing} extends this line to generation and grounding. As shown in Table~\ref{tab:finegrained_annotation_comparison}, these datasets usually represent an outfit as a coordinated set of items, together with item images, metadata, user interactions, or textual descriptions. Such supervision is well suited to matching-oriented tasks, where the main objective is to model relationships among multiple fashion items. What they usually lack is explicit supervision for holistic outfit interpretation from a human fashion perspective. In particular, they rarely annotate whether an outfit is stylistically coherent, seasonally appropriate, suitable for a given occasion, or supported by a clear styling rationale.

Overall, existing datasets typically emphasize either item-level perception, text-enhanced item understanding, or outfit/set-level matching structure. In contrast, \datasetname\ combines these aspects through professionally grounded annotations for both single fashion items and complete outfits, together with more realistic outfit scenarios involving accessories, layering, and richer multi-item composition. This design supports not only outfit-to-item grounding and outfit completion, but also expert-level outfit evaluation.

\section{Dataset Construction}

We design \datasetname\ as a fashion benchmark for expert-level understanding.
Unlike existing datasets that typically emphasize either item perception or outfit compatibility, \datasetname\ explicitly connects fine-grained item semantics with outfit-level contextual judgments. 
To this end, the dataset is organized into two aligned levels: \textit{item-level samples}, which capture the intrinsic properties of individual garments and accessories, and \textit{outfit-level samples}, which characterize the coordinated semantics of complete looks. 
This dual-level design enables \datasetname\ to support not only fashion perception, but also recommendation-oriented reasoning and LLM/MLLM-based fashion understanding.

\textbf{Data Collection and Curation}.
We collect the dataset from two large-scale real-world fashion e-commerce platforms, Taobao and Dewu\footnote{\url{https://www.taobao.com},  \url{https://www.dewu.com}}, covering menswear, womenswear, and childrenswear. To ensure practical relevance, sampling is performed across diverse clothing categories, style families, seasonal conditions, and usage scenarios. In addition to garments, the dataset includes functional accessories such as shoes, bags, hats, and scarves, since these items are essential to realistic outfit composition.

A key requirement during curation is complete outfit--item linkage: each retained outfit must be associated with valid records for all its constituent items. Starting from 2,239 raw outfits and 7,112 raw item records, we apply a multi-stage curation pipeline including duplicate removal, visual quality filtering, metadata normalization, and linkage verification. Images with severe occlusion, insufficient quality, or incomplete correspondence between outfit and item records are removed. For a small number of recoverable cases, AI-assisted restoration is applied and then manually verified by experts. After standardization, all images are converted to PNG format and resized to $512\times384$.

\textbf{Expert Annotation Pipeline.}
\datasetname\ is annotated by a \textit{\textbf{nine-member fashion-domain expert team}}, including one faculty member with over 20 years of experience in fashion design and eight master's students with formal training in fashion and apparel design. 
The team members specialize in areas such as digital fashion design, design theory, and apparel-oriented design research. 
To ensure consistency, the annotation process follows an iterative workflow in which annotation guidelines are continuously refined through cross-review and consistency verification, together with initial labeling and vocabulary normalization. Ambiguous semantic fields are normalized into predefined closed vocabularies, on which inter-annotator agreement measured by Cohen's $\kappa$~\cite{cohen1960coefficient} reaches 0.64, indicating substantial agreement~\cite{landis1977measurement}. Natural-language style descriptions are written under shared annotation guidelines and subject to cross-review among team members.
The annotation schema is designed to capture two complementary levels of fashion understanding:
\begin{itemize}[leftmargin=1.6em, itemsep=2pt, topsep=2pt, parsep=0pt, label=\diamondbullet]
    \item \textbf{Item-level semantics.} Item annotations go beyond coarse category labels and describe the physical, structural, and functional properties of fashion pieces. In particular, \emph{outline}, \emph{material}, \emph{pattern}, \emph{detail}, and \emph{layering role} provide useful cues for shape, warmth, design characteristics, and the functional role of each item within an outfit.
    \item \textbf{Outfit-level semantics.} Each outfit is annotated with a natural-language \emph{style description} written by fashion experts, together with \emph{season} and \emph{occasion} labels. These annotations capture holistic semantics that cannot be reduced to single item tags, including overall style, coordination logic, and contextual suitability.
\end{itemize}

\begin{figure}[tb]
    \centering
    \includegraphics[width=0.95\linewidth]{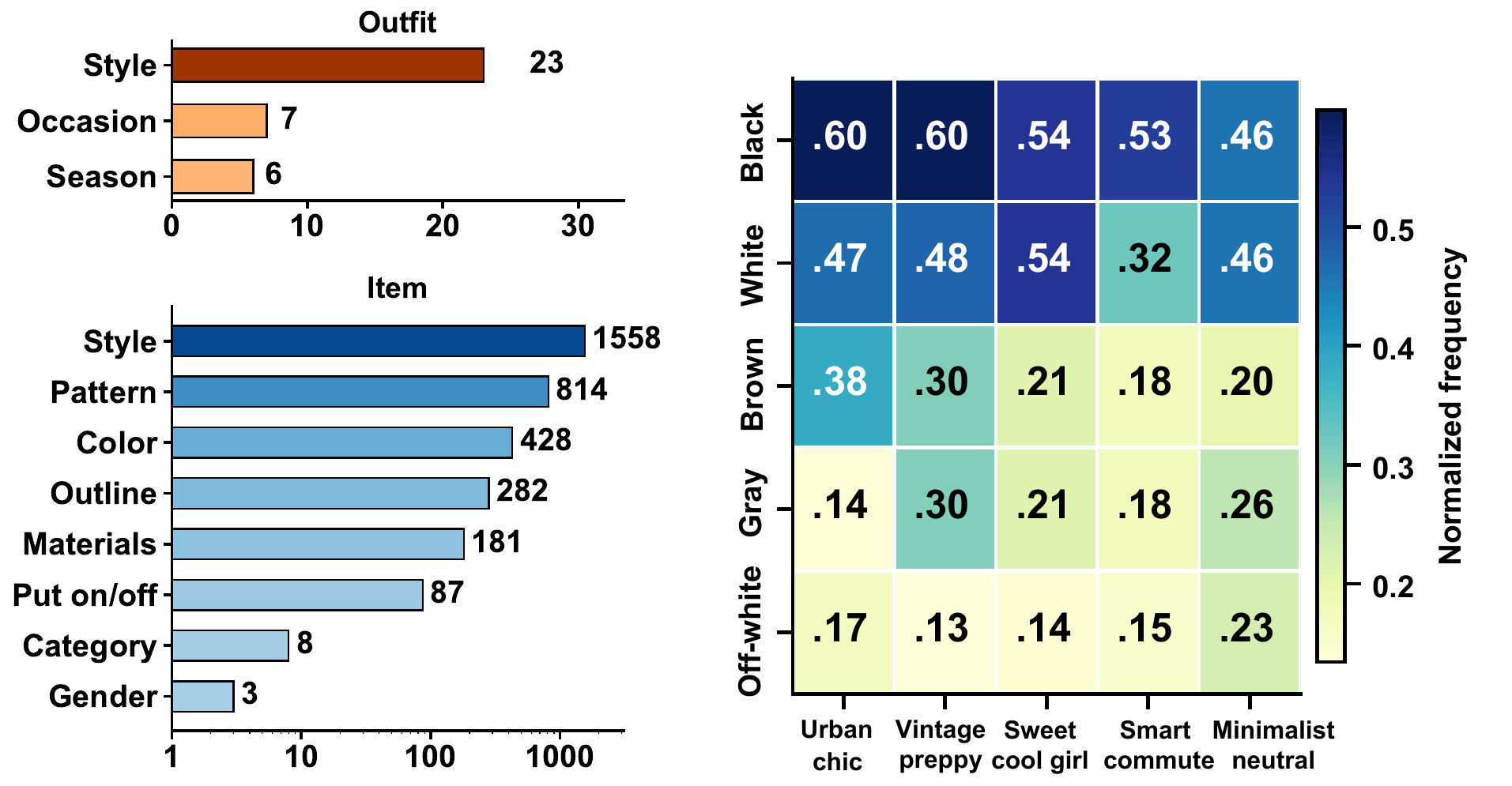}
    \captionsetup{skip=2pt}
    \caption{(\textit{Left}) Number of unique attribute values in \datasetname\ across item- and outfit-level annotations. (\textit{Right}) Normalized co-occurrence frequency between the top-5 most common colors and outfit styles.}
    \label{fig:color_style_heatmap}
    \vspace{-1.5em}
\end{figure}

\textbf{Dataset Statistics and Semantic Characteristics.}
The final dataset contains 1,000 outfit-level samples and 4,637 item-level samples, with broad coverage across menswear (300 outfits), womenswear (500 outfits), and childrenswear (200 outfits). On average, each outfit is associated with 4.6 items, including not only garments but also functional accessories such as \textit{shoes, bags, hats, scarves, and ties}, which better reflects the complexity of real-world outfit composition than simplified pairwise matching settings.

Figure~\ref{fig:color_style_heatmap} summarizes both the annotation coverage and semantic richness of \datasetname. The left panel shows that \datasetname\ contains diverse item- and outfit-level annotations, covering attributes such as style, pattern, color, outline, material, occasion, and season. This reflects the dual-level design of the dataset, which connects fine-grained item attributes with high-level outfit semantics.
The right panel shows that the relationship between visual attributes and outfit semantics is highly structured. For example, different colors exhibit clear style-specific preferences, indicating that the dataset captures meaningful correlations between visual appearance and fashion semantics. This suggests that \datasetname\ encodes fashion knowledge beyond low-level appearance cues.

More broadly, Figure~\ref{fig:color_style_heatmap} provides one simple example of the knowledge encoded in \datasetname. The dataset also contains many other fashion-specific regularities not shown here due to space limitations. Detailed examples can be found in our released dataset and annotations. Overall, these results highlight the value of \datasetname\ as a benchmark for knowledge-aware fashion understanding, style reasoning, and expert-aligned recommendation.

\textbf{Quality Verification.}
To further validate annotation quality, we conduct a verification study on a stratified 10\% sample of outfits. Each sampled outfit is decomposed into attribute-level review units covering both outfit and item annotations. Four fashion-domain experts then independently judge whether each attribute value is correct. Through majority voting, 91.3\% of the original annotations are endorsed as correct by the expert panel, confirming the reliability of the resulting dataset and our annotation pipeline. Due to space limitations, the detailed verification protocol will be released with our open-source code.

\textbf{Release Protocol.}
Following existing work~\cite{ni2019justifying,satar2025seeing}, we release source URLs, metadata, and annotations sufficient to support benchmark construction and evaluation. The dataset is intended exclusively for fashion understanding and recommendation research. Accordingly, the annotations are restricted to fashion-related semantics and explicitly exclude facial identity supervision. 
We also plan to continuously expand the dataset in future releases, while maintaining annotation quality and semantic reliability.
\vspace{-1em}
\section{Benchmark Task Formalization}
\label{sec:benchmark_task_formalization}

To comprehensively evaluate \datasetname, we design three benchmark tasks with increasing cognitive complexity. \textbf{Task \#1 (Outfit-to-item Grounding)} examines perceptual understanding by decomposing an outfit into constituent items via precise visual grounding. \textbf{Task \#2 (Outfit Completion)} advances to compositional analysis, requiring models to infer inter-item compatibility and select suitable items that complete a partial outfit. \textbf{Task \#3 (Outfit Evaluation)} demands holistic reasoning with expert-level judgment on stylistic coherence and contextual appropriateness.
In summary, these three tasks form a progressive evaluation hierarchy from perception through composition to expert-level evaluation, collectively covering the core practical needs of both \textbf{users} (\eg item search, outfit assembly) \cite{lin2020fashion,wang2019outfit} and\textbf{ fashion designers} (\eg styling assessment, collection curation) \cite{jeon2021fashionq,zhou2023fcboost,yang2020learning}.

Let $\mathcal{B}$ and $\mathcal{I}$ be the set of outfits and fashion items, respectively. Each outfit $b\in\mathcal{B}$ consists of multiple items from $\mathcal{I}$. Each outfit $b$ and item $i$ is associated with visual and textual information, denoted by $\mathbf{v}_b,\mathbf{v}_i$ and $\mathbf{t}_b,\mathbf{t}_i$, respectively. Based on these basic concepts, the proposed three benchmark tasks are defined as follows:

\noindent\textbf{Task \#1 (Outfit-to-item Grounding).} Given a full-look image $\mathbf{v}_b$ and the category of an item $c_i\in \mathbf{t}_i$, task \#1 requires the model to generate an image for item $i \in b$, such that the generated image recovers the corresponding ground-truth $\mathbf{v}_i$.

\noindent\textbf{Task \#2 (Outfit Completion).} Given a partial outfit $b^*$ obtained by randomly masking items from $b$, \ie $b^*\subsetneq b$, task \#2 requires the model to select appropriate items from the item pool $\mathcal{I}\setminus b^*$ to recover the complete outfit $b$.

\noindent\textbf{Task \#3 (Outfit Evaluation).} Given the item images of an outfit $b$, \ie $\{\mathbf{v}_i|i\in b\}$, together with the candidate sets of style $\mathcal{Y}$, occasion $\mathcal{O}$, and season $\mathcal{S}$, {extracted from the textual information of $\mathcal{I}$, \ie $\mathcal{Y,O,S}\subsetneq\{\mathbf{t}_i|i\in\mathcal{I}\}$}, task \#3 requires the model to predict the outfit's style, season, and occasion, and identifies the item that is stylistically inconsistent with the overall outfit.
\section{Experiments}
We conduct extensive experiments on \datasetname\ to evaluate the value of our FashionStylist as a benchmark and training resource for knowledge-aware fashion understanding. All experiments are conducted on \datasetname, which is split into training, validation, and test sets with a ratio of 7:1:2~\cite{sun2024adaptive, liu2025fine}. 
We conduct all experiments on a Linux server equipped with one NVIDIA A100 80GB GPU. 
All detailed experimental settings for each task are introduced separately below.

\begin{table}[t]
    \caption{Performance comparison on Task \#1, where $\uparrow$ ($\downarrow$) denotes higher (lower) is better, \textit{Improv.} denotes the relative improvement of \textsc{Sft} over \textsc{Default}, \textbf{bold} indicates the best within each model, and {\color{oursorange}\rule{0.8em}{0.8em}} indicates the best across all models.}
    \vspace{-0.1in}
    \label{tab:task1_results}
    \centering
    \setlength{\tabcolsep}{2pt}
    \small
    \resizebox{\columnwidth}{!}{%
    \begin{tabular}{llcccccc}
        \toprule
        \textbf{Model} & \textbf{Mode} & \textbf{R@10 ($\uparrow$)} & \textbf{N@10 ($\uparrow$)} & \textbf{PSNR ($\uparrow$)} & \textbf{SSIM ($\uparrow$)} & \textbf{FID ($\downarrow$)} & \textbf{KID ($\downarrow$)} \\
        \midrule
        \multirow{3}{*}{\makecell[l]{Flux.1-\\Kontext}}
         & \textsc{Default} & 0.0765 & 0.0456 & 11.574 & 0.7255 & 77.879 & \textbf{0.0120} \\
         & \textsc{Sft}     & \textbf{0.0890} & \textbf{0.0575} & \textbf{12.175} & \textbf{0.7351} & \textbf{71.709} & 0.0126 \\
        \cdashline{2-8}
         & \textit{Improv.} & +16.3\% & +26.1\% & +5.2\% & +1.3\% & +7.9\% & -5.0\% \\
        \midrule
        \multirow{3}{*}{\makecell[l]{Qwen-\\ImageEdit}}
         & \textsc{Default} & 0.1812 & 0.1138 & 10.544 & 0.6720 & 63.598 & 0.0104 \\
         & \textsc{Sft}     & \textbf{0.2254} & \textbf{0.1406} & \textbf{11.677} & \textbf{0.6782} & \textbf{45.505} & \textbf{0.0042} \\
        \cdashline{2-8}
         & \textit{Improv.} & +24.4\% & +23.6\% & +10.7\% & +0.9\% & +28.4\% & +59.6\% \\
        \midrule
        \multirow{3}{*}{\makecell[l]{LongCat-\\Turbo}}
         & \textsc{Default} & 0.2114 & 0.1445 & 12.096 & 0.7145 & 56.911 & 0.0092 \\
         & \textsc{Sft}     & \textbf{0.2460} & \textbf{0.1531} & \cellcolor{oursorange}\textbf{12.744} & \cellcolor{oursorange}\textbf{0.7489} & \textbf{55.668} & \textbf{0.0063} \\
        \cdashline{2-8}
         & \textit{Improv.} & +16.4\% & +6.0\% & +5.4\% & +4.8\% & +2.2\% & +31.5\% \\
        \midrule
        \makecell[l]{Nano\\Banana2}
         & \textsc{Default} & \cellcolor{oursorange}0.3304 & \cellcolor{oursorange}0.2240 & 12.204 & 0.6840 & \cellcolor{oursorange}42.944 & \cellcolor{oursorange}0.0035 \\
        \bottomrule
    \end{tabular}%
    }
\end{table}
\begin{figure}[tb]
    \centering
    \includegraphics[width=0.95\linewidth]{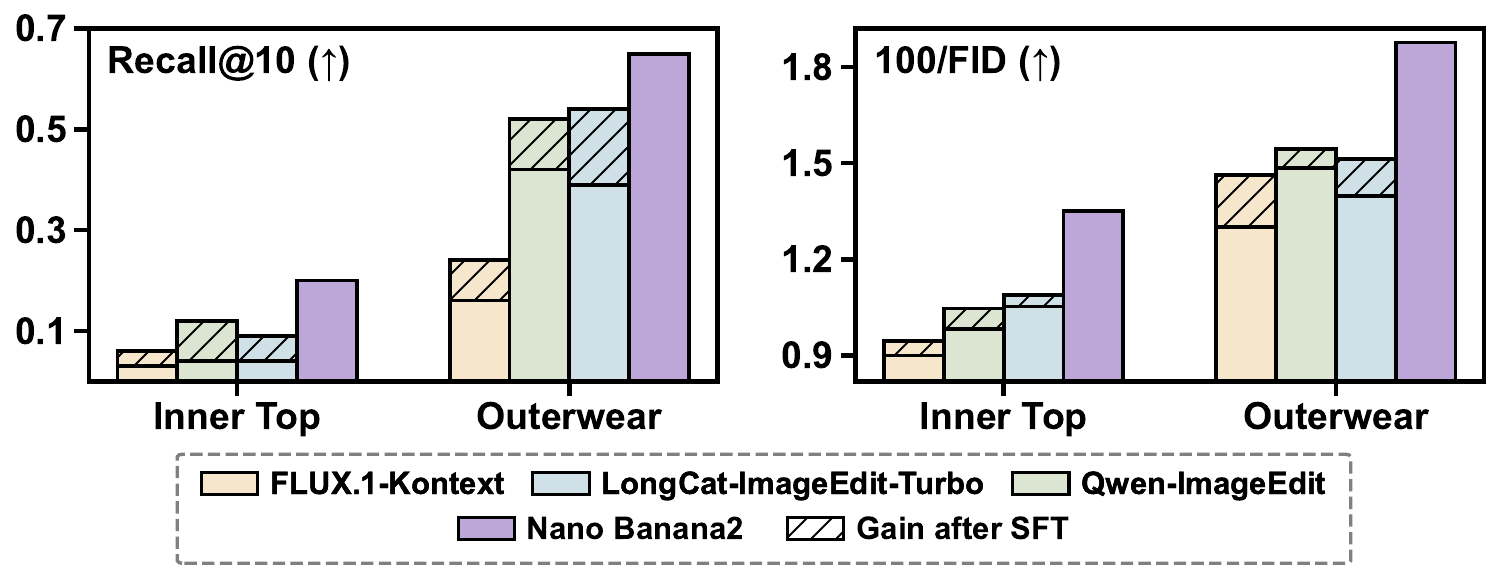}
    \vspace{-0.1in}
    \caption{Performance comparison across two representative item categories. We report 100/FID so that higher values consistently indicate better performance across all metrics.}
    \label{fig:category_compare}
    \vspace{-0.5cm}
\end{figure}
\subsection{Benchmarking on Task \#1}

\textbf{Baselines \& Evaluation Protocols.} 
We formulate Task \#1 as an image-editing task, where the input consists of an outfit image and a textual query specifying a target item category, and the model is required to extract the corresponding item from the outfit as a standalone image $\mathbf{v}_i$.
We select three advanced open-source image-editing models (FLUX.1 Kontext \cite{labs2025flux}, Qwen-ImageEdit \cite{wu2025qwen}, and LongCat-Turbo \cite{team2025longcat}) and one proprietary model (\ie Nano Banana2\footnote{https://ai.google.dev/gemini-api/docs/models/gemini-3.1-flash-image-preview}) as baselines. 
The generated images from these models can be used for direct comparison and retrieval.
In this case, we evaluate model performance from recommendation relevance (Recall@$k$ and NDCG@$k$), reconstruction fidelity (PSNR~\cite{kelecs2021psnr} and SSIM \cite{wang2004ssim}), and distributional quality (FID \cite{heusel2017fid}, and KID \cite{binkowski2018kid}).
Please refer to our supplemental material for more details about metric computation.

\textbf{Experimental Settings.} 
For the open-source models, we consider two settings: \textsc{Default} and supervised fine-tuning (\textsc{Sft}). 
The \textsc{Default} setting directly applies the pretrained model to Task \#1 without additional training, whereas \textsc{Sft} uses low-rank adaptation (LoRA) \cite{hu2022lora,dettmers2023qlora} to fine-tune the model on the training set and evaluate it on the test set. 

\textbf{Results Analysis.} The experimental results are reported in Table~\ref{tab:task1_results} and Figure~\ref{fig:category_compare}. \textbf{1)} \textsc{Sft} on our dataset consistently improves all three open-source models across nearly all metrics, showing that \textit{\textbf{the proposed dataset provides effective supervision for Task~\#1 and serves as a valuable resource for advancing fashion-oriented image editing}}. \textbf{2)} As shown in Figure~\ref{fig:category_compare}, all models perform markedly worse on \emph{Inner Top} than on \emph{Outerwear}, which we attribute to the complex layered styling in our dataset, where inner garments are often occluded by outer pieces. 
\textit{\textbf{This finding highlights the value of incorporating complex outfit compositions into benchmark construction}}, as recovering occluded items from holistic outfits remains challenging in realistic fashion scenarios. \textbf{3)} Although the proprietary NanoBanana2 remains to be a strong baseline, \textsc{Sft} narrows the gap between compact open-source models and the commercial system, \textit{\textbf{further underscoring the necessity and value of high-quality, fashion-specialized training data}}.




\begin{table}[t]
\caption{Performance comparison on Task \#2, where \textsc{Eke.} denotes ``Expert knowledge-enhanced'', \textit{Improv.} denotes the relative improvement of \textsc{Eke.} over \textsc{Default}.}
\vspace{-0.1in}
\label{tab:task2_results}
\centering
\footnotesize
\setlength{\tabcolsep}{2pt}
\renewcommand{\arraystretch}{1.1}
\resizebox{0.9\columnwidth}{!}{%
\begin{tabular}{cllcccc}
\toprule
\textbf{Type} & \textbf{Model} & \textbf{Mode} & \textbf{R@10 ($\uparrow$)} & \textbf{R@20 ($\uparrow$)} & \textbf{N@10 ($\uparrow$)} & \textbf{N@20 ($\uparrow$)} \\
\midrule
\multirow{9}{*}{\rotatebox{90}{Retrieval}}
& \multirow{3}{*}{POG}
& \textsc{Default} & 0.0022 & 0.0065 & 0.0007 & 0.0018 \\
& & \textsc{Eke.}    & \textbf{0.0054} & \textbf{0.0131} & \textbf{0.0028} & \textbf{0.0047} \\
\cdashline{3-7}
& & \textit{Improv.} & +145.5\% & +101.5\% & +300.0\% & +161.1\% \\
\cmidrule{2-7}

& \multirow{3}{*}{CIRP}
& \textsc{Default} & 0.0239 & 0.0423 & 0.0131 & 0.0175 \\
& & \textsc{Eke.}    & \textbf{0.0293} & \textbf{0.0456} & \textbf{0.0173} & \textbf{0.0213} \\
\cdashline{3-7}
& & \textit{Improv.} & +22.6\% & +7.8\% & +32.1\% & +21.7\% \\
\cmidrule{2-7}

& \multirow{3}{*}{CLHE}
& \textsc{Default} & 0.0304 & 0.0684 & 0.0119 & 0.0217 \\
& & \textsc{Eke.}    & \textbf{0.0380} & \textbf{0.0782} & \textbf{0.0193} & \textbf{0.0296} \\
\cdashline{3-7}
& & \textit{Improv.} & +25.0\% & +14.3\% & +62.2\% & +36.4\% \\
\midrule

\multirow{3}{*}{\rotatebox{90}{Gen.}}
& \multirow{3}{*}{DiFashion}
& \textsc{Default} & 0.0216 & 0.0519 & 0.0108 & 0.0196 \\
& & \textsc{Eke.}    & \cellcolor{oursorange}\textbf{0.0650} & \cellcolor{oursorange}\textbf{0.1000} & \cellcolor{oursorange}\textbf{0.0347} & \cellcolor{oursorange}\textbf{0.0441} \\
\cdashline{3-7}
& & \textit{Improv.} & +200.9\% & +92.7\% & +221.3\% & +125.0\% \\
\bottomrule
\end{tabular}%
}
\vspace{-0.1in}
\end{table}

\subsection{Benchmarking on Task \#2}

\textbf{Baselines \& Evaluation Protocols.} 
Task \#2 can be addressed using both retrieval-based and generative paradigms. 
To provide a comprehensive comparison, we include representative baselines from both paradigms, including retrieval-based methods (POG~\cite{chen2019pog}, CIRP~\cite{ma2024cirp}, and CLHE~\cite{ma2024leveraging}) and generative method (DiFashion~\cite{xu2024diffusion}). 
All selected baselines are compatible with multimodal inputs. 
Following prior work \cite{ma2024cirp,ma2024leveraging,xu2024diffusion}, we adopt Recall@$k$ and NDCG@$k$ ($k\in{10, 20}$) as evaluation metrics.

\textbf{Experimental Settings.} Following CIRP~\cite{ma2024cirp}, we adopt a leave-one-out strategy: one item is masked from each outfit and recovered from the item pool $\mathcal{I}\setminus b^*$. Inputs consist of images and texts of the items within one partial outfit under two settings: \textsc{Default} (item titles only) and \textsc{Eke.} (with fine-grained expert annotations such as style and material). Retrieval baselines share FashionCLIP~\cite{chia2204fashionclip} as the encoder. For the generative baseline, both the generated and pool images are encoded by ResNet-50, and the missing item is retrieved via cosine similarity to produce a ranked list.

\textbf{Results Analysis.} Table~\ref{tab:task2_results} reports the experimental results, which lead to the following observations: \textbf{1)} Across all models, the \textsc{Eke.} mode consistently outperforms \textsc{Default}, demonstrating that \textit{\textbf{the fine-grained expert knowledge annotated in our dataset provides effective supervision for outfit completion.}} \textbf{2)} The consistent gains brought by \textsc{Eke.} across retrieval-based methods show that \textit{\textbf{the expert annotations in our dataset offer complementary fashion knowledge that is not captured by standard item titles alone.}} In particular, the improvements on NDCG suggest that this enriched supervision mainly improves the ranking quality of compatible items. \textbf{3)} The substantial improvements under the \textsc{Eke.} setting further \textit{\textbf{indicate that our dataset is particularly valuable for knowledge-intensive fashion generation}}, where rich semantic cues are essential. This suggests that expertise-level knowledge in our dataset can directly guide the generation of compatible items through rich semantic cues, such as style, rather than only supporting candidate re-ranking.

\begin{table}[t]
\caption{Performance comparison on Task \#3, where {\itshape Improv.} indicates the relative gain of \textsc{Sft} over \textsc{Instruct}.}
\vspace{-0.1in}
\label{tab:task3_results}
\centering
\small
\resizebox{0.95\columnwidth}{!}{%
\begin{tabular}{llcccc}
\toprule
\textbf{Model} & \textbf{Mode} & \textbf{Style} ($\uparrow$) & \textbf{Season} ($\uparrow$) & \textbf{Occasion} ($\uparrow$) & \textbf{Mod.} ($\uparrow$) \\
\midrule
\multirow{4}{*}{Gemma3-4b}
& \textsc{Instruct}         & 0.1325 & 0.1950 & 0.2725 & 0.0100 \\
& \textsc{Think}            & 0.1425 & 0.2575 & 0.2425 & \textbf{0.1200} \\
& \textsc{Sft}              & \textbf{0.2500} & \textbf{0.4425} & \textbf{0.2875} & 0.0850 \\
\cdashline{2-6}
& \textit{Improv.} & +88.7\% & +126.9\% & +5.5\% & +750.0\% \\
\midrule
\multirow{4}{*}{Qwen3VL-4b}
& \textsc{Instruct}         & 0.1975 & 0.3250 & 0.2525 & 0.1150 \\
& \textsc{Think}            & 0.2125 & 0.3625 & 0.2250 & 0.1000 \\
& \textsc{Sft}              & \textbf{0.2650} & \textbf{0.4350} & \textbf{0.2875} & \textbf{0.1250} \\
\cdashline{2-6}
& \textit{Improv.} & {+34.2\%} & {+33.8\%} & {+13.9\%} & {+8.7\%} \\
\midrule
\multirow{4}{*}{Qwen2.5VL-7b}
& \textsc{Instruct}    & 0.1600 & 0.2600 & 0.2650 & 0.0000 \\
& \textsc{Think}       & 0.1550 & 0.2850 & 0.2525 & 0.0800 \\
& \textsc{Sft}         & \cellcolor{oursorange}\textbf{0.2850} & \cellcolor{oursorange}\textbf{0.5500} & \cellcolor{oursorange}\textbf{0.3000} & \textbf{0.1100} \\
\cdashline{2-6}
& \textit{Improv.} & {+78.1\%} & {+111.5\%} & {+13.2\%} & {--} \\
\midrule
\multirow{4}{*}{Qwen3VL-8b}
& \textsc{Instruct}         & 0.2400 & 0.3125 & 0.2550 & 0.1050 \\
& \textsc{Think}            & 0.2375 & 0.3525 & \textbf{0.2650} & 0.0650 \\
& \textsc{Sft}              & \textbf{0.2650} & \textbf{0.5200} & 0.2400 & \textbf{0.1400} \\
\cdashline{2-6}
& \textit{Improv.} & {+10.4\%} & {+66.4\%} & {-5.9\%} & {+33.3\%} \\
\midrule
Gemini 3.1-Pro & \textsc{Think} & 0.2175 & 0.3675 & 0.2525 & \cellcolor{oursorange}0.2600 \\
\bottomrule
\end{tabular}%
}
\vspace{-3pt}
\end{table}
\subsection{Benchmarking on Task \#3}

{\bf Baselines and Evaluation Protocols.} 
For Task \#3, we evaluate whether large models can assess outfit-level attributes and detect mismatched items based on the original \datasetname. 
To this end, we construct both intact outfits and corrupted outfits containing a style-incompatible item. The resulting evaluation set contains 2,000 samples, including 1,000 original outfits and 1,000 corrupted counterparts. We compare four representative open-source multimodal language large models (MLLMs), including Gemma3-4b \cite{gemmateam2025gemma3technicalreport}, Qwen2.5VL-7b \cite{bai2025qwen25vltechnicalreport}, and Qwen3VL-4b/8b \cite{bai2025qwen3vltechnicalreport}, together with one proprietary model, Gemini 3.1-Pro \cite{geminiteam2025geminifamilyhighlycapable}. 
We use {\it accuracy} as the evaluation metric for all sub-tasks. 
Specifically, given an outfit, the model is asked to predict its \textit{Style}, \textit{Season}, and \textit{Occasion}, and determine whether the outfit contains a style-mismatched item.
\textit{Mod.} denotes the accuracy of this mismatch detection task.


{\bf Experimental Settings.} The input consists of all item images in an outfit, along with candidate labels for style, season, and occasion. We design a prompt to instruct MLLMs to predict the outfit’s style, season, and occasion, and whether it contains a style-mismatched item. For the open-source baselines, we consider three settings: \textsc{Instruct}, \textsc{Think}, and \textsc{Sft}. In \textsc{Instruct}, the prompt is directly applied to the pre-trained MLLMs. In \textsc{Think}, the \texttt{<think>} token is appended to the prompt to activate model reasoning. In \textsc{Sft}, we fine-tune each MLLM using LoRA \cite{hu2022lora, dettmers2023qlora}. Due to space limitations, full prompts and implementation details are deferred to the supplementary material.

{\bf Results Analysis.} The experimental results are reported in Table~\ref{tab:task3_results}. We highlight three main observations. \textbf{1)} \textsc{Sft} improves performance in most cases, showing that \textit{\textbf{our dataset provides effective supervision for outfit understanding and enables even smaller open-source models to achieve strong performance}} on style, season, and occasion prediction. This suggests that existing MLLMs still lack sufficient fashion-domain knowledge, which can be effectively supplied by the expert annotations in our dataset through \textsc{Sft}. \textbf{2)} Activating reasoning (\textsc{Think}) yields consistent improvement only on season prediction, while its effect on the remaining dimensions is mixed. This indicates that \textit{\textbf{general-purpose reasoning is insufficient for several fashion understanding tasks}}, since season can often be inferred from coarse visual cues, whereas style, occasion, and mismatch detection rely more heavily on the fine-grained fashion knowledge provided by our dataset. \textbf{3)} Although Gemini achieves the best \textit{Mod.} score without fine-tuning, \textsc{Sft} still brings clear gains to open-source models on this task. This suggests that mismatch detection is more challenging than attribute prediction, as it requires cross-item comparison in addition to domain knowledge; nevertheless, \textit{\textbf{the gains from \textsc{Sft} further demonstrate the value of \datasetname\ for improving fine-grained outfit evaluation}}.

\section{Conclusion and Future Work}

In this paper, we present \datasetname, a publicly available benchmark for knowledge-aware fashion understanding and evaluation in realistic scenarios. Constructed through a fashion-expert annotation pipeline, \datasetname\ provides expert-grounded annotations at both the item and outfit levels, and supports three tasks: outfit-to-item grounding, outfit completion, and outfit evaluation. Our experiments show that current LLMs and MLLMs remain limited in expert-level fashion understanding, while \datasetname\ provides valuable supervision for improving both outfit recommendation and fine-grained outfit evaluation performance.

A current limitation of \datasetname\ is its moderate scale, which partly reflects the substantial time and cost of high-quality expert-driven annotation. In future work, we will continue to \textit{\textbf{update}} and \textit{\textbf{expand}} \datasetname\ in both scale and outfit diversity, especially toward more diverse, realistic, and complex dressing scenarios involving accessories, layering, and richer multi-item composition. We also plan to explore MLLM-assisted annotation to improve annotation efficiency and scalability while preserving high annotation quality and semantic reliability.

\newpage
\bibliographystyle{ACM-Reference-Format}
\bibliography{sample-base}

@String{Computing = "Computing" }

@String{Computer = "{IEEE} Computer" }

@String{Springer = "Springer-Verlag" }

@article{cohen1960coefficient,
  title={A coefficient of agreement for nominal scales},
  author={Cohen, Jacob},
  journal={Educational and psychological measurement},
  volume={20},
  number={1},
  pages={37--46},
  year={1960},
  publisher={Sage Publications Sage CA: Thousand Oaks, CA}
}

@article{landis1977measurement,
  title={The measurement of observer agreement for categorical data},
  author={Landis, J Richard and Koch, Gary G},
  journal={biometrics},
  pages={159--174},
  year={1977},
  publisher={JSTOR}
}

@inproceedings{satar2025seeing,
  title={Seeing culture: A benchmark for visual reasoning and grounding},
  author={Satar, Burak and Ma, Zhixin and Irawan, Patrick Amadeus and Mulyawan, Wilfried Ariel and Jiang, Jing and Lim, Ee-Peng and Ngo, Chong-Wah},
  booktitle={Proceedings of the 2025 Conference on Empirical Methods in Natural Language Processing},
  pages={22238--22254},
  year={2025}
}

@inproceedings{ni2019justifying,
  title={Justifying recommendations using distantly-labeled reviews and fine-grained aspects},
  author={Ni, Jianmo and Li, Jiacheng and McAuley, Julian},
  booktitle={Proceedings of the Conference on Empirical Methods in Natural Language Processing and the 9th International Joint Conference on Natural Language Processing},
  pages={188--197},
  year={2019}
}

@inproceedings{wang2019outfit,
  title={Outfit compatibility prediction and diagnosis with multi-layered comparison network},
  author={Wang, Xin and Wu, Bo and Zhong, Yueqi},
  booktitle={Proceedings of the 27th ACM International Conference on Multimedia},
  pages={329--337},
  year={2019}
}

@inproceedings{yang2020learning,
  title={Learning tuple compatibility for conditional outfit recommendation},
  author={Yang, Xuewen and Xie, Dongliang and Wang, Xin and Yuan, Jiangbo and Ding, Wanying and Yan, Pengyun},
  booktitle={Proceedings of the 28th ACM International Conference on Multimedia},
  pages={2636--2644},
  year={2020}
}

@inproceedings{zhou2023fcboost,
  title={Fcboost-net: A generative network for synthesizing multiple collocated outfits via fashion compatibility boosting},
  author={Zhou, Dongliang and Zhang, Haijun and Ma, Jianghong and Fan, Jicong and Zhang, Zhao},
  booktitle={Proceedings of the 31st ACM International Conference on Multimedia},
  pages={7881--7889},
  year={2023}
}

@inproceedings{su2021complementary,
  title={Complementary factorization towards outfit compatibility modeling},
  author={Su, Tianyu and Song, Xuemeng and Zheng, Na and Guan, Weili and Li, Yan and Nie, Liqiang},
  booktitle={Proceedings of the 29th ACM International Conference on Multimedia},
  pages={4073--4081},
  year={2021}
}

@inproceedings{jin2025genwardrobe,
  title={GenWardrobe: A fully generative system for travel fashion wardrobe construction},
  author={Jin, Peng and Wen, Yilin and Yu, Mingzhe and Ma, Yunshan and Zheng, Rong and Fan, Jin-tu and NGO, Chong Wah},
  booktitle={Proceedings of the 33rd ACM International Conference on Multimedia},
  pages={13540--13542},
  year={2025}
}

@inproceedings{ma2020knowledge,
  title={Knowledge enhanced neural fashion trend forecasting},
  author={Ma, Yunshan and Ding, Yujuan and Yang, Xun and Liao, Lizi and Wong, Wai Keung and Chua, Tat-Seng},
  booktitle={Proceedings of the 2020 International Conference on Multimedia Retrieval},
  pages={82--90},
  year={2020}
}

@inproceedings{guan2022personalized,
  title={Personalized fashion compatibility modeling via metapath-guided heterogeneous graph learning},
  author={Guan, Weili and Jiao, Fangkai and Song, Xuemeng and Wen, Haokun and Yeh, Chung-Hsing and Chang, Xiaojun},
  booktitle={Proceedings of the 45th international ACM SIGIR Conference on Research and Development in Information Retrieval},
  pages={482--491},
  year={2022}
}

@article{labs2025flux,
  title={FLUX. 1 Kontext: Flow matching for in-context image generation and editing in latent space},
  author={Labs, Black Forest and Batifol, Stephen and Blattmann, Andreas and Boesel, Frederic and Consul, Saksham and Diagne, Cyril and Dockhorn, Tim and English, Jack and English, Zion and Esser, Patrick and others},
  journal={arXiv preprint arXiv:2506.15742},
  year={2025}
}

@article{wu2025qwen,
  title={Qwen-image technical report},
  author={Wu, Chenfei and Li, Jiahao and Zhou, Jingren and Lin, Junyang and Gao, Kaiyuan and Yan, Kun and Yin, Sheng-ming and Bai, Shuai and Xu, Xiao and Chen, Yilei and others},
  journal={arXiv preprint arXiv:2508.02324},
  year={2025}
}

@article{team2025longcat,
  title={Longcat-image technical report},
  author={Team, Meituan LongCat and Ma, Hanghang and Tan, Haoxian and Huang, Jiale and Wu, Junqiang and He, Jun-Yan and Gao, Lishuai and Xiao, Songlin and Wei, Xiaoming and Ma, Xiaoqi and others},
  journal={arXiv preprint arXiv:2512.07584},
  year={2025}
}

@article{wang2004ssim,
  title={Image quality assessment: From error visibility to structural similarity},
  author={Wang, Zhou and Bovik, Alan C and Sheikh, Hamid R and Simoncelli, Eero P},
  journal={IEEE Transactions on Image Processing},
  volume={13},
  number={4},
  pages={600--612},
  year={2004},
  publisher={IEEE}
}

@article{heusel2017fid,
  title={GANs trained by a two time-scale update rule converge to a local nash equilibrium},
  author={Heusel, Martin and Ramsauer, Hubert and Unterthiner, Thomas and Nessler, Bernhard and Hochreiter, Sepp},
  journal={Advances in Neural Information Processing Systems},
  year={2017}
}

@article{binkowski2018kid,
  title={Demystifying mmd gans},
  author={Bi{\'n}kowski, Miko{\l}aj and Sutherland, Danica J and Arbel, Michael and Gretton, Arthur},
  journal={arXiv preprint arXiv:1801.01401},
  year={2018}
}

@article{kelecs2021psnr,
  title={On the computation of PSNR for a set of images or video},
  author={Kele{\c{s}}, Onur and Y{\i}lmaz, M Ak{\i}n and Tekalp, A Murat and Korkmaz, Cansu and Dogan, Zafer},
  journal={arXiv preprint arXiv:2104.14868},
  year={2021}
}

@inproceedings{lin2020fashion,
  title={Fashion outfit complementary item retrieval},
  author={Lin, Yen-Liang and Tran, Son and Davis, Larry S},
  booktitle={Proceedings of the IEEE/CVF Conference on Computer Vision and Pattern Recognition},
  pages={3311--3319},
  year={2020}
}

@inproceedings{jeon2021fashionq,
  title={FashionQ: An ai-driven creativity support tool for facilitating ideation in fashion design},
  author={Jeon, Youngseung and Jin, Seungwan and Shih, Patrick C and Han, Kyungsik},
  booktitle={Proceedings of the 2021 CHI Conference on Human Factors in Computing Systems},
  pages={1--18},
  year={2021}
}

@inproceedings{liu2025fine,
  title={Fine-tuning multimodal large language models for product bundling},
  author={Liu, Xiaohao and Wu, Jie and Tao, Zhulin and Ma, Yunshan and Wei, Yinwei and Chua, Tat-seng},
  booktitle={Proceedings of the 31st ACM SIGKDD Conference on Knowledge Discovery and Data Mining},
  pages={848--858},
  year={2025}
}

@inproceedings{sun2024adaptive,
  title={Adaptive in-context learning with large language models for bundle generation},
  author={Sun, Zhu and Feng, Kaidong and Yang, Jie and Qu, Xinghua and Fang, Hui and Ong, Yew-Soon and Liu, Wenyuan},
  booktitle={Proceedings of the 47th International ACM SIGIR Conference on Research and Development in Information Retrieval},
  pages={966--976},
  year={2024}
}

@inproceedings{ma2024cirp,
  title={Cirp: Cross-item relational pre-training for multimodal product bundling},
  author={Ma, Yunshan and He, Yingzhi and Zhong, Wenjun and Wang, Xiang and Zimmermann, Roger and Chua, Tat-Seng},
  booktitle={Proceedings of the 32nd ACM International Conference on Multimedia},
  pages={9641--9649},
  year={2024}
}

@inproceedings{ma2024leveraging,
  title={Leveraging multimodal features and item-level user feedback for bundle construction},
  author={Ma, Yunshan and Liu, Xiaohao and Wei, Yinwei and Tao, Zhulin and Wang, Xiang and Chua, Tat-Seng},
  booktitle={Proceedings of the 17th ACM International Conference on Web Search and Data Mining},
  pages={510--519},
  year={2024}
}

@article{chia2204fashionclip,
  title={Contrastive language and vision learning of general fashion concepts},
  author={Chia, Patrick John and Attanasio, Giuseppe and Bianchi, Federico and Terragni, Silvia and Magalhaes, Ana Rita and Goncalves, Diogo and Greco, Ciro and Tagliabue, Jacopo},
  journal={Scientific Reports},
  volume={12},
  number={1},
  pages={18958},
  year={2022},
  publisher={Nature Publishing Group UK London}
}

@misc{gemmateam2025gemma3technicalreport,
      title={Gemma 3 technical report}, 
      author={Gemma Team},
      year={2025},
      eprint={2503.19786},
      archivePrefix={arXiv},
      primaryClass={cs.CL},
      url={https://arxiv.org/abs/2503.19786}, 
}

@misc{bai2025qwen25vltechnicalreport,
      title={Qwen2.5-VL technical report}, 
      author={Shuai Bai and Keqin Chen and Xuejing Liu and Jialin Wang and Wenbin Ge and Sibo Song and Kai Dang and Peng Wang and others},
      year={2025},
      eprint={2502.13923},
      archivePrefix={arXiv},
      primaryClass={cs.CV},
      url={https://arxiv.org/abs/2502.13923}, 
}

@misc{bai2025qwen3vltechnicalreport,
      title={Qwen3-VL technical report}, 
      author={Shuai Bai and Yuxuan Cai and Ruizhe Chen and Keqin Chen and Xionghui Chen and Zesen Cheng and Lianghao Deng and Wei Ding and others},
      year={2025},
      eprint={2511.21631},
      archivePrefix={arXiv},
      primaryClass={cs.CV},
      url={https://arxiv.org/abs/2511.21631}, 
}

@article{geminiteam2025geminifamilyhighlycapable,
  title={Gemini: A family of highly capable multimodal models. arXiv 2023},
  author={Team, Gemini and Anil, Rohan and Borgeaud, Sebastian and Alayrac, Jean-Baptiste and Yu, Jiahui and Soricut, Radu and Schalkwyk, Johan and Dai, Andrew M and Hauth, Anja and Millican, Katie and others},
  journal={arXiv preprint arXiv:2312.11805},
  year={2024}
}

@inproceedings{hu2022lora,
  title={LoRA: Low-rank adaptation of large language models},
  author={Hu, Edward J and Shen, Yelong and Wallis, Phillip and Allen-Zhu, Zeyuan and Li, Yuanzhi and Wang, Shean and Wang, Lu and Chen, Weizhu},
  booktitle={Proceedings of the International Conference on Learning Representations (ICLR)},
  year={2022}
}

@article{dettmers2023qlora,
  title={Qlora: Efficient finetuning of quantized llms},
  author={Dettmers, Tim and Pagnoni, Artidoro and Holtzman, Ari and Zettlemoyer, Luke},
  journal={Advances in Neural Information Processing Systems},
  pages={10088--10115},
  year={2023}
}

@inproceedings{liu2016DeepFashion,
  author={Liu, Ziwei and Luo, Ping and Qiu, Shi and Wang, Xiaogang and Tang, Xiaoou},
  booktitle={2016 IEEE Conference on Computer Vision and Pattern Recognition (CVPR)}, 
  title={DeepFashion: Powering robust clothes recognition and retrieval with rich annotations}, 
  year={2016},
  volume={},
  number={},
  pages={1096-1104},
  doi={10.1109/CVPR.2016.124}}

@inproceedings{ge2019deepfashion2,
  title={Deepfashion2: A versatile benchmark for detection, pose estimation, segmentation and re-identification of clothing images},
  author={Ge, Yuying and Zhang, Ruimao and Wang, Xiaogang and Tang, Xiaoou and Luo, Ping},
  booktitle={Proceedings of the IEEE/CVF Conference on Computer Vision and Pattern Recognition},
  pages={5337--5345},
  year={2019}
}

@inproceedings{chen2019pog,
  title={POG: personalized outfit generation for fashion recommendation at Alibaba iFashion},
  author={Chen, Wen and Huang, Pipei and Xu, Jiaming and Guo, Xin and Guo, Cheng and Sun, Fei and Li, Chao and Pfadler, Andreas and Zhao, Huan and Zhao, Binqiang},
  booktitle={Proceedings of the 25th ACM SIGKDD International Conference on Knowledge Discovery \& Data Mining},
  pages={2662--2670},
  year={2019}
}

@inproceedings{jia2020fashionpedia,
  title={Fashionpedia: Ontology, segmentation, and an attribute localization dataset},
  author={Jia, Menglin and Shi, Mengyun and Sirotenko, Mikhail and Cui, Yin and Cardie, Claire and Hariharan, Bharath and Adam, Hartwig and Belongie, Serge},
  booktitle={European Conference on Computer Vision},
  pages={316--332},
  year={2020},
  organization={Springer}
}

@inproceedings{lu2019learning,
  title={Learning binary code for personalized fashion recommendation},
  author={Lu, Zhi and Hu, Yang and Jiang, Yunchao and Chen, Yan and Zeng, Bing},
  booktitle={Proceedings of the IEEE/CVF Conference on Computer Vision and Pattern Recognition},
  pages={10562--10570},
  year={2019}
}

@article{rostamzadeh2018fashion,
  title={Fashion-gen: The generative fashion dataset and challenge},
  author={Rostamzadeh, Negar and Hosseini, Seyedarian and Boquet, Thomas and Stokowiec, Wojciech and Zhang, Ying and Jauvin, Christian and Pal, Chris},
  journal={arXiv preprint arXiv:1806.08317},
  year={2018}
}

@inproceedings{wu2021fashion,
  title={Fashion IQ: A new dataset towards retrieving images by natural language feedback},
  author={Wu, Hui and Gao, Yupeng and Guo, Xiaoxiao and Al-Halah, Ziad and Rennie, Steven and Grauman, Kristen and Feris, Rogerio},
  booktitle={Proceedings of the IEEE/CVF Conference on Computer Vision and Pattern Recognition},
  pages={11307--11317},
  year={2021}
}

@inproceedings{song2019gp,
  title={GP-BPR: Personalized compatibility modeling for clothing matching},
  author={Song, Xuemeng and Han, Xianjing and Li, Yunkai and Chen, Jingyuan and Xu, Xin-Shun and Nie, Liqiang},
  booktitle={Proceedings of the 27th ACM international Conference on Multimedia},
  pages={320--328},
  year={2019}
}

@article{deshmukh2024dressing,
  title={Dressing the imagination: A Dataset for AI-powered translation of text into fashion outfits and a novel NeRA adapter for enhanced feature adaptation},
  author={Deshmukh, Gayatri and De, Somsubhra and Sehgal, Chirag and Gupta, Jishu Sen and Mittal, Sparsh},
  journal={arXiv preprint arXiv:2411.13901},
  year={2024}
}

@inproceedings{zheng2021personalized,
  title={Personalized fashion recommendation from personal social media data: An item-to-set metric learning approach},
  author={Zheng, Haitian and Wu, Kefei and Park, Jong-Hwi and Zhu, Wei and Luo, Jiebo},
  booktitle={2021 IEEE International Conference on Big Data },
  pages={5014--5023},
  year={2021},
  organization={IEEE}
}

@article{cheng2021fashion,
  title={Fashion meets computer vision: A survey},
  author={Cheng, Wen-Huang and Song, Sijie and Chen, Chieh-Yun and Hidayati, Shintami Chusnul and Liu, Jiaying},
  journal={ACM Computing Surveys},
  volume={54},
  number={4},
  pages={1--41},
  year={2021},
  publisher={ACM New York, NY, USA}
}

@article{ding2023computational,
  title={Computational technologies for fashion recommendation: A survey},
  author={Ding, Yujuan and Lai, Zhihui and Mok, PY and Chua, Tat-Seng},
  journal={ACM Computing Surveys},
  volume={56},
  number={5},
  pages={1--45},
  year={2023},
  publisher={ACM New York, NY}
}

@article{shi2025generative,
  title={Generative AI in fashion: Overview},
  author={Shi, Wenda and Wong, Waikeung and Zou, Xingxing},
  journal={ACM Transactions on Intelligent Systems and Technology},
  volume={16},
  number={4},
  pages={1--73},
  year={2025},
  publisher={ACM New York, NY}
}

@inproceedings{xu2024diffusion,
  title={Diffusion models for generative outfit recommendation},
  author={Xu, Yiyan and Wang, Wenjie and Feng, Fuli and Ma, Yunshan and Zhang, Jizhi and He, Xiangnan},
  booktitle={Proceedings of the 47th International ACM SIGIR Conference on Research and Development in Information Retrieval},
  pages={1350--1359},
  year={2024}
}

@inproceedings{zhao2024unifashion,
  title={Unifashion: A unified vision-language model for multimodal fashion retrieval and generation},
  author={Zhao, Xiangyu and Zhang, Yuehan and Zhang, Wenlong and Wu, Xiao-Ming},
  booktitle={Proceedings of the 2024 Conference on Empirical Methods in Natural Language Processing},
  pages={1490--1507},
  year={2024}
}

@inproceedings{liao2018knowledge,
  title={Knowledge-aware multimodal fashion chatbot},
  author={Liao, Lizi and Zhou, You and Ma, Yunshan and Hong, Richang and Chua, Tat-seng},
  booktitle={Proceedings of the 26th ACM International Conference on Multimedia},
  pages={1265--1266},
  year={2018}
}

@inproceedings{choi2021viton,
  title={Viton-hd: High-resolution virtual try-on via misalignment-aware normalization},
  author={Choi, Seunghwan and Park, Sunghyun and Lee, Minsoo and Choo, Jaegul},
  booktitle={Proceedings of the IEEE/CVF Conference on Computer Vision and Pattern Recognition},
  pages={14131--14140},
  year={2021}
}

@inproceedings{morelli2022dress,
  title={Dress code: High-resolution multi-category virtual try-on},
  author={Morelli, Davide and Fincato, Matteo and Cornia, Marcella and Landi, Federico and Cesari, Fabio and Cucchiara, Rita},
  booktitle={Proceedings of the IEEE/CVF Conference on Computer Vision and Pattern Recognition},
  pages={2231--2235},
  year={2022}
}

\end{document}